\documentclass{article}
\usepackage{microtype}
\usepackage{graphicx}
\usepackage{subfigure}
\usepackage{booktabs} 
\usepackage{multirow,bm}
\usepackage{tikz}
\usepackage{threeparttable}
\usepackage{tikz-cd,mathtools}
\usepackage{mathtools}
\usetikzlibrary{arrows, matrix} 
\usepackage{xcolor}
\definecolor{darkblue}{rgb}{0.0, 0.0, 0.55}
\definecolor{darkred}{rgb}{0.55, 0.0, 0.0}
\usepackage{hyperref}
\usepackage{comment}
\usepackage[accepted]{icml2024}  % 如果你是在icml2024模板下编译
\usepackage{amsmath}
\usepackage{amssymb}
\usepackage{mathtools}
\usepackage{amsthm}
\usepackage{multirow}
\usepackage[capitalize,noabbrev]{cleveref}
\usepackage{enumitem}
\usepackage{microtype}
\usepackage{graphicx}
\usepackage{subfigure}
\usepackage{booktabs} 
\usepackage{multirow,bm}
\usepackage{tikz}
\usepackage{threeparttable}
\usepackage{tikz-cd,mathtools}
\usepackage{mathtools}
\usetikzlibrary{arrows, matrix} 
\usepackage{xcolor}
\definecolor{darkblue}{rgb}{0.0, 0.0, 0.55}
\definecolor{darkred}{rgb}{0.55, 0.0, 0.0}
\usepackage{hyperref}
\usepackage{wrapfig}
\usepackage[utf8]{inputenc} % allow utf-8 input
\usepackage[T1]{fontenc}    % use 8-bit T1 fonts
\usepackage{hyperref}       % hyperlinks
\usepackage{url}            % simple URL typesetting
\usepackage{booktabs}       % professional-quality tables
\usepackage{amsfonts}       % blackboard math symbols
\usepackage{nicefrac}       % compact symbols for 1/2, etc.
\usepackage{microtype}      % microtypography
\usepackage{xcolor}         % colors
\usepackage{tikz}

\usepackage{booktabs}
\usepackage{subfigure}
\usepackage{enumitem}
\usepackage{rotating}
\usepackage[utf8]{inputenc}

\usepackage[ruled,vlined]{algorithm2e}

\usepackage{amsmath,amssymb,amsfonts}
\newtheorem{theorem}{Theorem}

\usepackage{multirow}
\usepackage{booktabs,bm}

\usepackage{colortbl}
\usepackage{multirow}
\usepackage{booktabs}
\usepackage{adjustbox}
\usepackage{amsmath}
\usepackage{mathtools}
\usepackage[utf8]{inputenc}
\usepackage{hyperref}
\usepackage{comment}
\usepackage{wrapfig}
\usepackage{amsmath}
\usepackage{graphicx}
\usepackage{subcaption}
\usepackage{multirow}
\usepackage{booktabs}
\newcommand{\showcomments}{yes}
\newcommand\hao[1]{
    \ifthenelse{\equal{\showcomments}{yes}}{{\color{red} Hao: #1}}{\ignorespaces}
}
\newcommand\xingjian[1]{
    \ifthenelse{\equal{\showcomments}{yes}}{{\color{blue} Xingjian: #1}}{\ignorespaces}
}
\newcommand\weiyan[1]{
    \ifthenelse{\equal{\showcomments}{yes}}{{\color{green} Weiyan: #1}}{\ignorespaces}
}

\usepackage{amsmath}         % 数学公式支持
\usepackage{enumitem}        % 列表定制
\setlist[itemize]{leftmargin=*, align=left} % 全局列表左对齐设置

\usepackage{enumitem}    % 用于自定义列表
\usepackage{amsmath}     % 增强数学排版
\usepackage{natbib}      % 处理引用
\usepackage{graphicx}    % 插入图片
\usepackage{comment}     % 使用 comment 环境
\usepackage{wrapfig}     % 环绕图像

\def\method{BeamVQ}% The \icmltitle you define below is probably too long as a header.

\icmltitlerunning{\method: Beam Search with Vector Quantization to Mitigate Data Scarcity in Physical Spatiotemporal Forecasting}

\begin{document}

\twocolumn[
\icmltitle{\method: Beam Search with Vector Quantization to Mitigate Data Scarcity in Physical Spatiotemporal Forecasting}




\begin{icmlauthorlist}
\icmlauthor{Weiyan Wang}{tx}
\icmlauthor{Xingjian Shi}{Boson}
\icmlauthor{Ruiqi Shu}{thu1}
\icmlauthor{Yuan Gao}{thu1}
\icmlauthor{Rui Ray Chen}{thu2}
\icmlauthor{Kun Wang}{ntu}
\icmlauthor{Fan Xu}{ustc}
\icmlauthor{Jinbao Xue}{tx}
\icmlauthor{Shuaipeng Li}{tx}
\icmlauthor{Yangyu Tao}{tx}
\icmlauthor{Di Wang}{tx}
\icmlauthor{Hao Wu}{tx,thu1,ustc}
\icmlauthor{Xiaomeng Huang}{thu1}
\end{icmlauthorlist}
\vskip -0.03in
\icmlaffiliation{tx}{TEG, Tencent}
\icmlaffiliation{Boson}{Boson AI}
\icmlaffiliation{thu1}{Department of Earth System Science, Ministry of Education Key Laboratory for Earth System Modeling, Institute for Global Change Studies, Tsinghua University}
\icmlaffiliation{ustc}{Department and Computer and Science, University of Science and Technology of China}
\icmlaffiliation{thu2}{Institute for Interdisciplinary Information Sciences, Tsinghua University}
\icmlaffiliation{ntu}{School of Computer Science and Engineering, Nanyang Technological University}

\icmlcorrespondingauthor{Xiaomeng Huang}{hxm@tsinghua.edu.cn}

\icmlkeywords{Machine Learning, ICML}

\vskip 0.3in
]

\printAffiliationsAndNotice{}  

\begin{abstract}
In practice,  physical spatiotemporal forecasting can suffer from data scarcity, because collecting large-scale data is non-trivial, especially for extreme events. 
Hence, we propose \method{}, a novel probabilistic framework to realize iterative self-training with new self-ensemble strategies, 
achieving better physical consistency and generalization on extreme events. 
Following any base forecasting model, 
we can encode its deterministic outputs into a latent space and retrieve multiple codebook entries to generate probabilistic outputs. 
Then \method{} extends the beam search from discrete spaces to the continuous state spaces in this field.
We can further employ domain-specific metrics (e.g., Critical Success Index for extreme events) to filter out the top-k candidates and develop the new self-ensemble strategy by combining the high-quality candidates. 
The self-ensemble can not only improve the inference quality and robustness but also iteratively augment the training datasets during continuous self-training. 
Consequently, \method{} realizes the exploration of rare but critical phenomena beyond the original dataset. 
Comprehensive experiments on different benchmarks and backbones show that \method{} consistently reduces forecasting MSE (up to 39\%), enhancing extreme events detection and proving its effectiveness in handling data scarcity. Our codes are available at~\url{https://github.com/easylearningscores/BeamVQ}.

% 在气象预报、流体模拟以及基于偏微分方程（PDE）的多物理系统模型中，数据稀缺下的时空预测仍然是一个关键挑战。本文提出了\method{}，一个统一的框架，旨在同时解决标注数据有限以及在确保物理一致性的前提下捕捉极端事件的难题。首先，我们训练了一个确定性的基础模型，从小规模数据中学习主要动力学。随后，通过Top-K 向量量化变分自编码器（VQ-VAE）对基础模型的输出进行增强，该模块将确定性预测编码到潜在空间，并检索多个码本条目以生成多样化且物理上合理的重构结果。一个新颖的联合优化过程利用领域特定的指标（例如关键成功指数）引导基础模型向更准确且对极端事件敏感的预测方向优化。在推理阶段，我们采用束搜索策略，维持多个候选轨迹并通过指标感知评分进行迭代剪枝，从而在探索罕见但关键现象与利用最可能的系统轨迹之间实现平衡。在多个气象和流体流动基准数据集上的大量实验表明，\method{}显著提升了预测精度，增强了对极端状态的检测能力，并保持了物理合理性，证明了其在数据稀缺场景下进行时空预测的优越性。

\end{abstract}

\section{Introduction}

In physical spatiotemporal forecasting (e.g., meteorological forecasting~\cite{bi2023accurate, lam2022graphcast}, fluid simulation~\cite{wu2024prometheus, wupure}, and various multiphysics system models~\cite{li2020fourier, wu2024neural}),
%driven by partial differential equations (PDEs)~\cite{li2020fourier, wu2024neural}), 
researchers typically need to capture physical patterns and predict extreme events,
such as heavy rainfall due to severe convective weather~\cite{ravuri2021skilful,doswell2001severe}, marine heatwave~\cite{frolicher2018marine}, and intense turbulence~\cite{moisy2004geometry}). 
However, they suffer from the fundamental problem of data scarcity to ensure physical consistency and accurately predict extreme events.
Collecting large-scale and high-resolution physical data can be expensive and even infeasible.
Consequently, limited training data can prevent data driven models~\cite{sun2020surrogate, zhu2019physics} like physics-informed neural networks~\cite{raissi2019physics} from generalizing well, even though they have adopted physical laws as prior knowledge.
Furthermore, extreme events occur infrequently in nature, making their labeled data quite sparse and imbalanced throughout the entire data set.
Therefore, data-driven methods usually fail to capture these low-probability phenomena.
%On one hand, for physical processes with clear equations or prior laws, lacking sufficient and high-quality observational data often prevents good generalization when using Physics-Informed Neural Networks (PINNs)~\cite{raissi2019physics} or other physics-constrained models~\cite{sun2020surrogate, zhu2019physics} for large-scale high-dimensional predictions. 
%On the other hand, extreme events occur infrequently in nature, making them difficult and costly to observe, which results in even more limited labeled data. \textit{This limitation reduces the ability of conventional data-driven models or ensemble forecasting methods to capture these low-probability phenomena.}
\begin{figure}[t]
  \centering
  \includegraphics[width=1\linewidth]{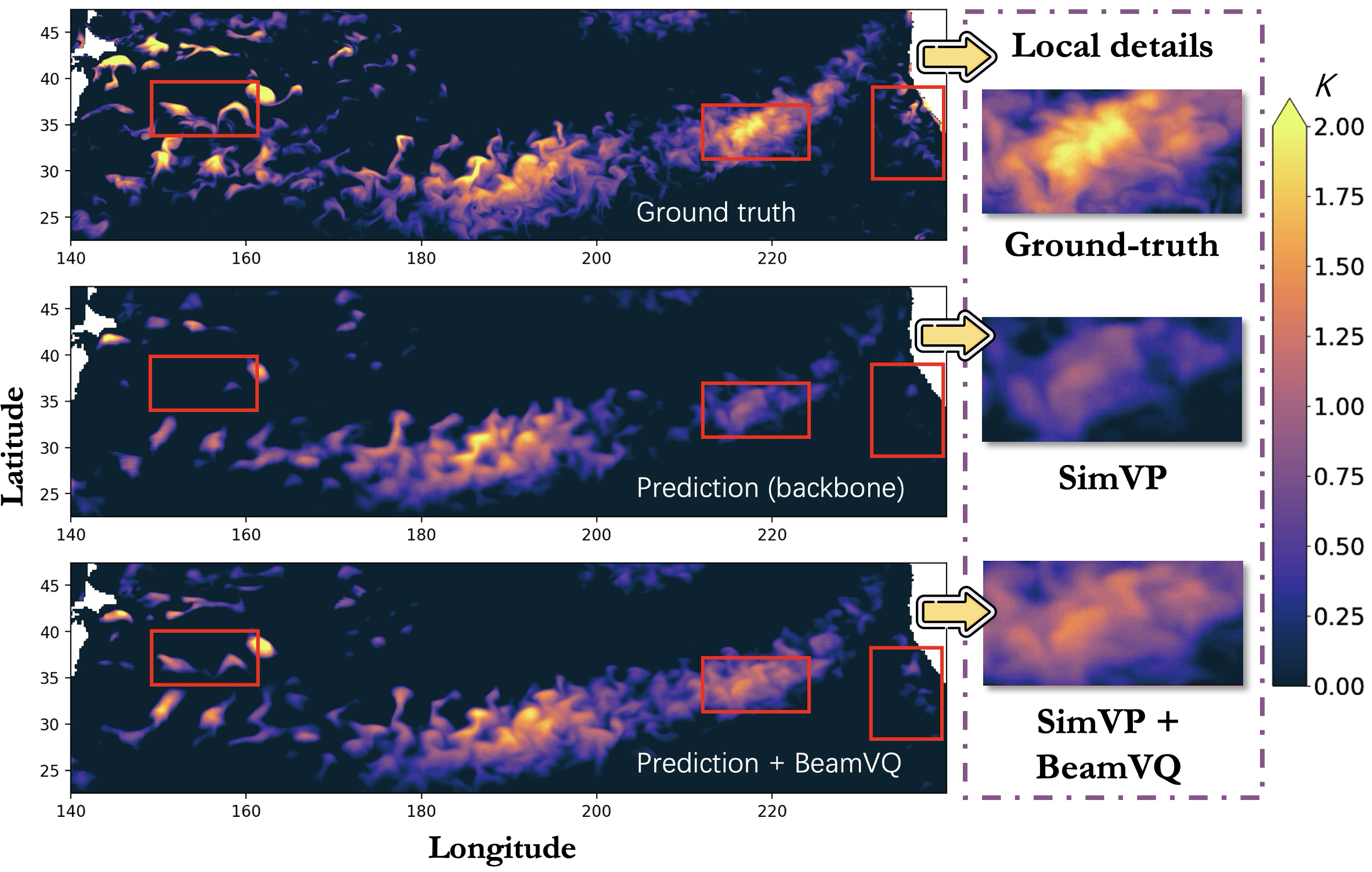}
  \caption{The visualization of extreme marine heatwave events shows that BeamVQ enhances Backbone models and improves their ability to capture extreme events. Detailed experimental results are provided in the experiments section.}
  \label{fig:intro}
\end{figure}

Existing studies on physical spatiotemporal forecasting belong to two categories, namely \textit{Numerical Methods} and \textit{Data-driven Methods}.
Traditional numerical methods like finite difference and finite element simulate future changes by solving physical equations~\cite{jouvet2009numerical, rogallo1984numerical, orszag1974numerical}. Although these numerical methods can be consistent with fundamental physical principles, they not only suffer incredibly expensive computations but also can be infeasible, if we cannot fully understand the underlying mechanism of complex or rare physical events~\cite{takamoto2022pdebench}.
Moreover, they are sensitive to the input disturbance (aka. the Butterfly Effect~\cite{lorenz1972predictability}), so they usually perturb initial conditions with different random noises to make multiple predictions for ensemble forecasting~\cite{LEUTBECHER20083515, karlbauer2024advancing}, resulting in even higher time costs.
    %Similarly,  physics-constrained models like PINN~\cite{raissi2019physics, li2021physics,hansen2023learning} leverage physical equations as additional loss terms to improve physical consistency and generalization, but they can only work on limited problems that have simplified equations and fixed boundary conditions.

Driven by large-scale data, deep learning has emerged as the revolutionary approach for complex physical systems~\cite{shi2015convolutional, gao2022earthformer, tan2022simvp}.
    Some methods attempt to combine physical knowledge with model training for better physical consistency and generalization~\cite{long2018pde,greydanus2019hamiltonian,cranmer2020lagrangian}. 
    For example, PhyDNet~\cite{guen2020disentangling} and FNO~\cite{li2020fourier} add physics-inspired operators into the deep networks.
    Some works also introduce generative settings into extreme weather simulations, but they require numerical simulations to generate enough artificial data~\cite{zhang2023skilful, ravuri2021skilful}.
    Some works like PINN~\cite{raissi2019physics, li2021physics,hansen2023learning} leverage physical equations as additional loss regularization, which only works on specific problems with simplified equations and fixed boundary conditions. 
    PreDiff~\cite{gao2024prediff} trains a latent diffusion model with guidance from a physics-informed energy function. 
    Since these works manipulate physical prior as soft constraints to optimize the statistical metrics across the existing data,
    they rely on large-scale and high-quality data that is non-trivial to collect.
    %and some methods can even require numerical simulations to generate enough artificial data~\cite{raissi2019physics, ravuri2021skilful, zhang2023skilful}. 
    Worse still, extreme events are always a small proportion of the full set, leading to poor prediction.

Some works in other files have explored various techniques to alleviate the data shortage, but they focus on the classification tasks and exploit domain characters.
For example, Computer Vision (CV) develops data augmentation like Mixup~\cite{zhang2017mixup}, which mixes different images and their labels to generate new samples. 
Natural Language Processing (NLP) conducts self-training to make use of extra unlabeled data with pseudo labels~\cite{du2020self}.
CV also widely adopts self-ensemble like EMA models~\cite{wang2022self} to improve the robustness.
However, physical spatiotemporal forecasting cannot directly adopt these domain-specific methods designed for classification tasks. More details about all related works are in Appendix~\ref{sec:related_work}.

% todo: need revise! by weiyan
%To fundamentally alleviate the two major challenges of \textit{physical constraints} and \textit{extreme events} (both of which face modeling difficulties due to \textbf{data scarcity}), this paper proposes~\method{},
To mitigate the problem of \underline{\textit{data scarcity}} in physical spatiotemporal forecasting, 
we propose Beam Search with Vector Quantization (\textbf{\method{}}) to improve physic consistency and generalization on extreme events.
At its core, it extends the beam search from discrete states typically in NLP to continuous state spaces of this field, enabling the self-ensemble of top-quality outputs for iterative self-training. 
%It specifically focuses on spatio-temporal prediction tasks under "data scarcity" and emphasizes the collaborative design of data augmentation and model self-training. 
Specifically, BeamVQ as a plugin, we can follow previous works to train a base spatiotemporal predictor to generate deterministic outputs. And we construct a variational quantization framework with a vector code book to realize the vector quantization (VQ) mechanism, which discretizes the continuous output spaces of the base prediction. Therefore, we can conduct beam searches through the time steps in physical spatiotemporal forecasting, in a similar way to NLP sentence generation. Through the beam search, we can filter out top-k good-quality candidates with any metrics (even the non-differentiable ones that cannot be directly optimized), leading to better exploration of possible future evolution paths.  Then \method{} develops a new self-ensemble strategy by combining all the top-k candidates. Besides improving the final predicting quality and robustness, the self-ensemble of top candidates can work as additional pseudo samples to iteratively augment the data set for continuous self-training, 
resulting in better physical consistency and generalization even on extreme events.
For example, Figure~\ref{fig:intro} demonstrates our capability in extreme marine heatwave events, whose frequency ranges from one to three annual events~\cite{oliver2018longer}. 

In summary, \method{} has the following main contributions:

\textit{\textbf{Novel Methodology}}. We introduce the \method{} framework, which discretizes outputs via Vector Quantization. By combining Beam Search and self-ensemble strategies, it efficiently explores possible future evolution paths. This approach can significantly enhances the capture of extreme events and increases prediction diversity.

\textit{\textbf{New Training Strategy}}. During the self-training phase, we incorporate "pseudo-labeled" samples from beam search into the training data and iteratively update the model. This process effectively compensates for the lack of real labels and indirectly optimizes any metrics for better physical consistency.

\textit{\textbf{Consistent Improvement}}. We conduct systematic evaluations on multiple datasets, including meteorological, fluid, and PDE simulations, and on different backbone networks such as CNN, RNN, and Transformer. \method{} reduces the average MSE by $18.97\%\sim39.08\%$, showing consistent and significant improvements in accuracy, extreme event capture, and physical plausibility, demonstrating our effectiveness in mitigating data scarcity.

\section{Method}
\textbf{Problem Definition.} We investigate spatiotemporal prediction tasks spanning meteorological forecasting~\cite{bi2023accurate}, computational fluid dynamics~\cite{wu2024prometheus}, and PDE-based systems~\cite{wu2024neural}. The observational data is structured as a 4D tensor $\mathbf{X} \in \mathbb{R}^{T \times C \times H \times W}$, where $T$ denotes temporal steps, $C$ represents physical variables (temperature, pressure, velocity fields), and $(H,W)$ specify spatial resolution. Our objective is to learn a parametric mapping $f_\Theta: \mathbf{X}_t \mapsto \hat{\mathbf{Y}}_{t+1}$ that predicts subsequent system states from historical sequences $\mathbf{X}_t = \{\mathbf{X}_1, ..., \mathbf{X}_t\}$. The parameters $\Theta$ are optimized through maximum likelihood estimation:
\begin{equation}
\Theta^* = \arg\max_{\Theta} \sum_{i=1}^T \log P(\mathbf{Y}_{t+1}^i | \mathbf{X}_t^i; \Theta)
\end{equation}
where $P(\mathbf{Y}_{t+1}^i | \mathbf{X}_t^i; \Theta)$ defines the predictive distribution. The optimized model enables multi-step forecasting via iterative rollout $\hat{\mathbf{Y}}_{t+k} = f_\Theta(\{\mathbf{X}_t, \hat{\mathbf{Y}}_{t+1}, ..., \hat{\mathbf{Y}}_{t+k-1}\})$, crucial for applications requiring temporal extrapolation in climate modeling~\cite{bi2023accurate}, combustion dynamics~\cite{anonymous2024openck}, and fluid simulations~\cite{wupure}.

\begin{figure*}[t]
\centering
\includegraphics[width=1\textwidth]{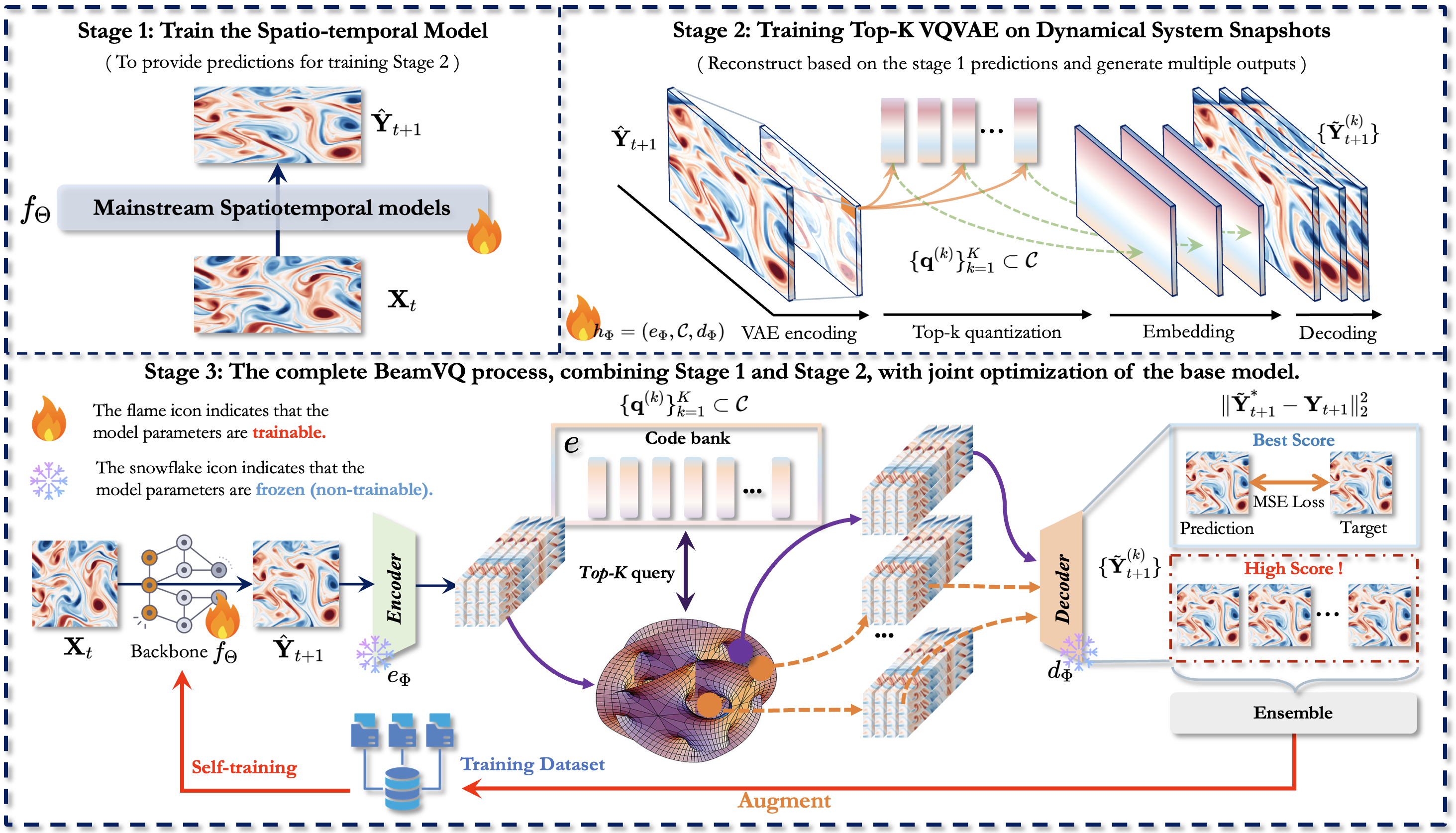}
\caption{\textbf{Architecture Overview of~\method{}.}  
(a) \textbf{Stage $1$: Base Model Training}: A deterministic predictor (FNO/ViT/ConvLSTM) learns single-step mappings $\mathbf{X}_t \xrightarrow{f_{\Theta_f}} \hat{\mathbf{Y}}_{t+1}$ via MSE minimization.  
(b) \textbf{Stage $2$: Top-K VQ-VAE}: Latent code $\mathbf{z}$ from encoder $e_{\Phi_h}$ is quantized to $K$ nearest codebook vectors $\{\mathbf{q}^{(k)}\}$, decoded to diverse predictions $\{\tilde{\mathbf{Y}}_{t+1}^{(k)}\}$.  
(c) \textbf{Joint Optimization}: The optimal reconstruction $\tilde{\mathbf{Y}}_{t+1}^*$ (selected by metric $M$) guides base model refinement, while top-$K'$ ensemble $\bar{\mathbf{Y}}_{t+1}$ enables self-training.} 
\label{fig:Idea_main} 
\end{figure*}

\textbf{Architecture Overview.} Our framework comprises three core stages of progressive refinement, as shown in Figure \ref{fig:Idea_main}. Initially, we train a base spatiotemporal predictor $f_\Theta$ that processes historical observations $\mathbf{X}_t \in \mathbb{R}^{1 \times C \times H \times W}$ (single-step training) to generate next-step predictions $\hat{\mathbf{Y}}_{t+1} = f_\Theta(\mathbf{X}_t)$. Subsequently, we develop a Top-K VQ-VAE $h_{\phi}$ through codebook-based pretraining, where the encoder $e_\phi$ maps $\hat{\mathbf{Y}}_{t+1}$ to latent code $\mathbf{z}$, quantized via K-nearest codebook vectors $\{\mathbf{q}^{(k)}\}_{k=1}^K \subset \mathcal{C}$, followed by decoder $d_\phi$ reconstruction to yield diverse outputs $\{\tilde{\mathbf{Y}}_{t+1}^{(k)}\}$. During joint optimization, we employ a non-differentiable metric $M$ (e.g., Critical Success Index~\cite{gao2022earthformer}) to select the optimal reconstruction $\tilde{\mathbf{Y}}_{t+1}^*$, then minimize $\|\tilde{\mathbf{Y}}_{t+1}^* - \mathbf{Y}_{t+1}\|_2^2$ to refine $f_\Theta$, while augmenting training data with ensemble averages of top-$K'$ candidates. For multi-step inference, beam search~\cite{steinbiss1994improvements} maintains $K$ trajectory candidates per step, progressively selecting optimal sequences through metric-guided pruning.

\subsection{Stage $1$: Pre-training the Base Prediction Model}
We first develop a foundational predictor $f_{\Theta}$ that learns deterministic spatiotemporal dynamics from observational data. The model ingests input tensors $\mathbf{X}_t \in \mathbb{R}^{1 \times C \times H \times W}$ (single-step temporal context during training) and generates predictions $\hat{\mathbf{Y}}_{t+1} = f_{\Theta}(\mathbf{X}_t)$ through parametric mapping $f_{\Theta}: \mathbb{R}^{C \times H \times W} \to \mathbb{R}^{C \times H \times W}$. Architectural implementations are task-adaptive: Fourier Neural Operators (FNO)~\cite{li2020fourier} for spectral systems governed by PDEs, Vision Transformers~\cite{dosovitskiy2020image} for global dependency modeling, or ConvLSTM~\cite{shi2015convolutional} networks for local spatiotemporal correlations. The parameters $\Theta$ are learned by minimizing the reconstruction error:
\begin{equation}
\mathcal{L}_{\text{base}} = \mathbb{E} \left\| \hat{\mathbf{Y}}_{t+1} - \mathbf{Y}_{t+1} \right\|_2^2
\end{equation}
where the expectation is over training pairs $(\mathbf{X}_t, \mathbf{Y}_{t+1})$. Optimization employs gradient-based methods (Adam)~\cite{kingma2014adam} with learning rate annealing, ensuring stable convergence. This stage establishes a strong deterministic prior that captures dominant physical patterns - for instance, FNO architectures learn Green's functions in Fourier space for fluid dynamics, while transformer variants attend to long-range atmospheric interactions. The trained $f_{\Theta^*}$ provides initial point estimates for subsequent uncertainty-aware refinement.

\subsection{Stage $2$: Top-K VQ-VAE Pre-training}  
We construct a variational quantization framework $h_\Phi = (e_\Phi, \mathcal{C}, d_\Phi)$ to learn diverse reconstructions from deterministic predictions. Given the base model output $\hat{\mathbf{Y}}_{t+1}$, the encoder $e_\Phi$ projects it to latent code:
\begin{equation}
\mathbf{z} = e_\Phi(\hat{\mathbf{Y}}_{t+1}) \in \mathbb{R}^{d_z}
\end{equation}
A codebook $\mathcal{C} = \{\mathbf{c}_i\}_{i=1}^N \subset \mathbb{R}^{d_z}$ with $N$ entries enables Top-K vector quantization:
\begin{equation}
\mathbf{q}^{(k)} = \mathop{\text{argmin}}_{\mathbf{c} \in \mathcal{C}} \|\mathbf{z} - \mathbf{c}\|_2^2 \quad \text{for } 1 \leq k \leq K
\end{equation}
The decoder $d_\Phi$ reconstructs $K$ variants through parallel decoding:
\begin{equation}
\tilde{\mathbf{Y}}_{t+1}^{(k)} = d_\Phi(\mathbf{q}^{(k)}), \quad 1 \leq k \leq K
\end{equation}
The training objective combines three components:
\begin{equation}\small
\mathcal{L}_{\text{VQ}} = \underbrace{\|\tilde{\mathbf{Y}}_{t+1}^{(1)} - \mathbf{Y}_{t+1}\|_2^2}_{\text{Reconstruction}} + \underbrace{\|\text{sg}[\mathbf{z}] - \mathbf{q}^{(1)}\|_2^2}_{\text{Codebook Learning}} + \beta\underbrace{\|\mathbf{z} - \text{sg}[\mathbf{q}^{(1)}]\|_2^2}_{\text{Commitment}}
\end{equation}
where $\text{sg}[\cdot]$ denotes stop-gradient operator and $\beta$ balances latent commitment. This design ensures:  
\textbf{(1)} Accurate primary mode reconstruction via $\tilde{\mathbf{Y}}_{t+1}^{(1)}$ optimization;  
\textbf{(2)} Codebook diversity preservation through Top-K retrieval;  
\textbf{(3)} Stable encoder-codebook alignment via commitment loss.

We conducted several experiments to verify the effect of selecting different $K$. And we use an optimization to explain how to choose $K$ to achieve the best performance.

\begin{theorem}
    The best selection of $K$ is determined by the numerical solution of the following optimization problem
    \begin{align}
    & \arg\min_{\boldsymbol{\pi}}\quad h(\boldsymbol{\pi},\mathbb{T}):=\boldsymbol{\pi}^\top A_\mathbb{T}\boldsymbol{\pi}, \\
    & \text{subject to} \quad
        \begin{cases}
            \sum_{i = 1}^N\pi_iT_i\leq\alpha, \\
            \sum_{i = 1}^N\pi_i = 1, \\
            0\leq\pi_i\leq m^{-1}, \quad 1\leq i\leq N.
        \end{cases}
    \end{align}
    where $\pi_i$ is the sampling probability of the augmented data
\end{theorem}

For details of the proof, please refer to Appendix~\ref{sec:proof}.

The pre-trained $h_{\Phi^*}$ establishes a structured latent manifold~\cite{han2018structured} that encapsulates both predictive fidelity and uncertainty, which will be leveraged in Stage $3$ for probabilistic refinement.

\subsection{Stage $3$: Joint Optimization}  
We develop a dual-phase optimization framework to refine the base predictor $f_\Theta$ using the frozen Top-K VQ-VAE $h_\Phi$. The process iterates between:
\begin{equation}
\hat{\mathbf{Y}}_{t+1} = f_\Theta(\mathbf{X}_t), \quad \{\tilde{\mathbf{Y}}_{t+1}^{(k)}\}_{k=1}^K = h_\Phi(\hat{\mathbf{Y}}_{t+1})
\end{equation}
where $h_\Phi$ remains fixed with $\Phi = \Phi^*$ from Stage $2$. A domain-specific metric $M$ (e.g., Critical Success Index) evaluates each reconstruction:
\begin{equation}
s^{(k)} = M(\tilde{\mathbf{Y}}_{t+1}^{(k)}, \mathbf{Y}_{t+1}), \quad k \in [1,K]
\end{equation}
\textbf{Optimization Cycle is as follows:}

1. \textit{Optimal Guidance}: Select the highest-scoring variant  
\begin{equation}
k^* = \mathop{\arg\max}\limits_{k} s^{(k)}, \quad \mathcal{L}_{\text{guide}} = \|\tilde{\mathbf{Y}}_{t+1}^{(k^*)} - \mathbf{Y}_{t+1}\|_2^2
\end{equation}

2. \textit{Ensemble Distillation}: Aggregate top-$K'$ candidates  
\begin{equation}
\bar{\mathbf{Y}}_{t+1} = \frac{1}{K'}\sum_{k=1}^{K'} \tilde{\mathbf{Y}}_{t+1}^{(k_{\text{top}})}
\end{equation}
where $k_{\text{top}}$ indexes the $K'$ highest $s^{(k)}$.

3. \textit{Parameter Update}:  
\begin{equation}
\Theta \leftarrow \Theta - \eta\nabla_\Theta(\mathcal{L}_{\text{guide}} + \lambda\|\hat{\mathbf{Y}}_{t+1} - \bar{\mathbf{Y}}_{t+1}\|_2^2)
\end{equation}
The frozen VQ-VAE acts as an uncertainty-aware teacher: $\mathcal{L}_{\text{guide}}$ aligns predictions with metric-optimal reconstructions. Ensemble distillation $\bar{\mathbf{Y}}_{t+1}$ mitigates exposure bias through data augmentation. Hyperparameter $\lambda$ balances direct supervision and distributional smoothing  

% This dual optimization propagates gradient signals through both deterministic predictions and codebook-induced variations, enhancing the base model's capacity to capture:  
% i) Dominant dynamical modes via $\mathcal{L}_{\text{guide}}$  
% ii) Capturing distributional diversity via ensemble regularization 
% iii) Extreme event signatures through metric-aware selection  

% The algorithm terminates when validation $M$ saturates, yielding $f_{\Theta^\dagger}$ with improved physical consistency and uncertainty quantification.

\subsection{Inference Stage with Beam Search}
We propose a novel beam search protocol that synergizes the base predictor $f_\Theta$ with the diversity-generating VQ-VAE $h_\Phi$. The algorithm maintains $B$ candidate trajectories to balance exploration (via codebook variations) and exploitation (through metric-guided selection), crucial for chaotic systems where small deviations amplify exponentially. The procedure (Algorithm~\ref{alg:beam_search}) operates in three phases:

\textbf{Initialization}: Generate $K$ initial variants from $\mathbf{X}_t$ using the VQ-VAE's decoding diversity

\textbf{Iterative Rollout}: At each step $n$, expand $B$ candidates into $B\times K$ possibilities using the codebook

\textbf{Trajectory Selection}: Retain top-$B$ paths based on accumulated scores $s_n^{(b,k)} = \sum_{m=t+1}^n S(\tilde{\mathbf{Y}}_m^{(b,k)})$

\begin{algorithm}[h]
\SetAlgoLined
\DontPrintSemicolon
\caption{Beam Search with Codebook Variations}\label{alg:beam_search}
\KwIn{Initial state $\mathbf{X}_t \in \mathbb{R}^{C\times H\times W}$, beam width $B$, horizon $N$}
\KwOut{Optimal trajectory $\{\tilde{\mathbf{Y}}_{t+1}^*,...,\tilde{\mathbf{Y}}_{t+N}^*\}$}
  
\emph{// Phase 1: Initialization}\;
$\hat{\mathbf{Y}}_{t+1} \leftarrow f_\Theta(\mathbf{X}_t)$\;
$\{\tilde{\mathbf{Y}}_{t+1}^{(k)}\}_{k=1}^K \leftarrow h_\Phi(\hat{\mathbf{Y}}_{t+1})$\;
$\mathcal{B}_{t+1} \leftarrow \text{Top-}B\left( \{ (\tilde{\mathbf{Y}}_{t+1}^{(k)}, S(\tilde{\mathbf{Y}}_{t+1}^{(k)})) \}_{k=1}^K \right)$\;

\emph{// Phase 2: Iterative Rollout}\;
\For{$n \leftarrow t+2$ \textbf{to} $t+N$}{
    $\mathcal{C}_n \leftarrow \emptyset$\;
    \ForEach{beam $b \in \mathcal{B}_{n-1}$}{
        $\hat{\mathbf{Y}}_n^{(b)} \leftarrow f_\Theta(\tilde{\mathbf{Y}}_{n-1}^{(b)})$\; 
        $\{\tilde{\mathbf{Y}}_n^{(b,k)}\}_{k=1}^K \leftarrow h_\Phi(\hat{\mathbf{Y}}_n^{(b)})$\;
        \For{$k\leftarrow 1$ \textbf{to} $K$}{
            $s_n^{(b,k)} \leftarrow s_{n-1}^{(b)} + \alpha^{n-t}S(\tilde{\mathbf{Y}}_n^{(b,k)})$\; 
            $\mathcal{C}_n \leftarrow \mathcal{C}_n \cup \{(\{\tilde{\mathbf{Y}}\text{ sequence}\}, s_n^{(b,k)})\}$\;
        }
    }
    $\mathcal{B}_n \leftarrow \mathop{\arg\max}\limits_{\substack{\mathcal{S}\subset\mathcal{C}_n \\ |\mathcal{S}|=B}} \sum_{(\cdot,s)\in\mathcal{S}} s$\;
}

\emph{// Phase 3: Terminal Selection}\;
$\{\tilde{\mathbf{Y}}^*\} \leftarrow \mathop{\arg\max}\limits_{(\mathcal{Y},s)\in\mathcal{B}_{t+N}} s$\;

\Return $\{\tilde{\mathbf{Y}}_{t+1}^*,...,\tilde{\mathbf{Y}}_{t+N}^*\}$\;
\end{algorithm}

\paragraph{Key Enhancements}  
Our beam search extends conventional approaches through:

\begin{itemize}
\item \textbf{Codebook-Driven Diversity}: The VQ-VAE generates $K$ physically-plausible variations at each step, avoiding mode collapse in chaotic systems. For weather prediction, this captures alternative storm trajectories that single-point estimates miss.

\item \textbf{Exponential Score Discounting~\cite{wang2024nuwadynamics}}: The term $\alpha^{n-t}$ ($\alpha \in (0,1]$) in the scoring function prioritizes recent accuracy, crucial for non-stationary processes. This implements:
\begin{equation}
s_n^{(b,k)} = \sum_{m=t+1}^n \alpha^{n-m} S(\tilde{\mathbf{Y}}_m^{(b,k)})
\end{equation}
\item \textbf{Dynamic Beam Pruning}: The selection operator $\arg\max_{\mathcal{S}}$ solves a knapsack-like optimization to maximize total score while maintaining beam width $B$. This is equivalent to:
\begin{equation}
\mathcal{B}_n = \underset{\mathcal{S}}{\text{maximize}} \sum_{(\cdot,s)\in\mathcal{S}} s \quad \text{s.t. } |\mathcal{S}| \leq B
\end{equation}
\end{itemize}
The whole algorithm of the proposed~\method{} is summarized in Algorithm~\ref{alg:framework}.
\begin{algorithm}[h]
\SetAlgoLined
\DontPrintSemicolon
\caption{Unified Framework of \method{}}\label{alg:framework}
\KwIn{Historical observations $\mathbf{X}_t$, prediction horizon $N$}
\KwOut{Optimal trajectory $\{\tilde{\mathbf{Y}}_{t+1}^*, ..., \tilde{\mathbf{Y}}_{t+N}^*\}$}

\emph{// Stage $1$: Base Model Training}\;
Initialize predictor $f_\Theta$ \;
Train $f_\Theta$ via $\min_\Theta \|\hat{\mathbf{Y}}_{t+1} - \mathbf{Y}_{t+1}\|_2^2$\;

\emph{// Stage $2$: VQ-VAE Codebook Learning}\;
Learn encoder $e_\Phi$, decoder $d_\Phi$, codebook $\mathcal{C}$\;
Generate $K$ variants $\{\tilde{\mathbf{Y}}_{t+1}^{(k)}\}$ per prediction\;

\emph{// Stage $3$: Joint Optimization}\;
\While{not converged}{
    Generate candidates $\{\tilde{\mathbf{Y}}_{t+1}^{(k)}\}$ via $h_\Phi$\;
    Select best candidate $\tilde{\mathbf{Y}}_{t+1}^*$ using metric $M$\; 
    Update $f_\Theta$ with $\tilde{\mathbf{Y}}_{t+1}^*$ and top-$K'$ ensemble\;
}

\emph{// Stage $4$: Beam Search Inference}\;
Initialize beam with top-$B$ candidates\;
\For{$n = t+1$ \textbf{to} $t+N$}{
    Expand each beam with $K$ codebook variants\;
    Keep top-$B$ trajectories by accumulated scores\;
}
\Return Best trajectory from final beam\;
\end{algorithm}
\section{Experiment}
\label{sec:experiment}
\begin{table*}[htbp]
  \centering
  \begin{sc}
  \caption{Performance comparison of various models with and without the \method{} method across five benchmark tests (SWE(u), RBC, NSE, Prometheus, SEVIR), using MSE as the evaluation metric. We bold-case the entries with lower MSE. ``Improvement'' represents the average percentage improvement in MSE achieved with \method{}.}
    \label{tab:mainres}
    \resizebox{\linewidth}{!}{%
      \begin{tabular}{l|cc|cc|cc|cc|cc}
        \toprule
        \multirow{4}{*}{Model} & \multicolumn{10}{c}{Benchmarks}  \\
        \cmidrule(lr){2-11}
        & \multicolumn{2}{c}{SWE (u)} & \multicolumn{2}{c}{RBC} & \multicolumn{2}{c}{NSE} &  \multicolumn{2}{c}{Prometheus} &  \multicolumn{2}{c}{SEVIR}   \\
        \cmidrule(lr){2-11}
        & Ori & + \method{} & Ori & + \method{} & Ori & + \method{} & Ori & + \method{}& Ori & + \method{} \\
        \midrule
        ResNet &0.0076 & \textbf{0.0033} & 0.1599 &\textbf{0.1283} & 0.2330 &\textbf{0.1663}  & 0.2356 &\textbf{0.1987} &  0.0671&  \textbf{0.0542} \\
        ConvLSTM &0.0024 &\textbf{0.0016}  &0.2726  & \textbf{0.0868} & 0.4094& \textbf{0.1277} & 0.0732 &\textbf{0.0533}  & 0.1757 & \textbf{0.1283}  \\
        Earthformer & 0.0135&\textbf{0.0093}  & 0.1273 &\textbf{0.1093} &1.8720  &  \textbf{0.1202} &0.2765  &\textbf{0.2001}  &0.0982  & \textbf{0.0521}  \\
        SimVP-v2 & 0.0013&\textbf{0.0010}  & 0.1234 & \textbf{0.1087} & 0.1238 &\textbf{0.1022}  & 0.1238 &\textbf{0.0921}  &0.0063&  \textbf{0.0032}\\
        TAU &0.0046 & \textbf{0.0031} & 0.1221 &\textbf{0.0965}  & 0.1205 &\textbf{0.1017} & 0.1201 &\textbf{0.0899}& 0.0059 &  \textbf{0.0029} \\
        Earthfarseer &0.0075 &\textbf{0.0059}  & 0.1454 &\textbf{0.1023}  & 0.1138 &\textbf{0.0987}  & 0.1176 &\textbf{0.1092}&  0.0065  &  \textbf{0.0021}  \\
        FNO & 0.0031&  \textbf{0.0024}& 0.1235 & \textbf{0.1053} & 0.2237 & \textbf{0.1005} & 0.3472 & \textbf{0.2275} & 0.0783 &  \textbf{0.0436} \\
        NMO &0.0021 &\textbf{0.0004}  &0.1123  & \textbf{0.1092} & 0.1007 & \textbf{0.0886} &0.0982  &\textbf{0.0475}  & 0.0045 & \textbf{0.0029}  \\
        CNO & 0.0146&  \textbf{0.0016}& 0.1327 & \textbf{0.1086} &0.2188 & \textbf{0.1483} &0.1097  &  \textbf{0.0254}&0.0056 &  \textbf{0.0053}  \\
        FourcastNet &0.0065 &\textbf{0.0061}  & 0.0671 & \textbf{0.0219} & 0.1794 &\textbf{0.1424}  &0.0987  &\textbf{0.0542}  & 0.0721 &\textbf{0.0652}   \\
        \midrule
         Improvement(\%)& \multicolumn{2}{c|}{+39.08$\%$}  &  \multicolumn{2}{c|}{+18.97$\%$}   &   \multicolumn{2}{c|}{+35.83$\%$}   &  \multicolumn{2}{c|}{+33.65$\%$}   &  \multicolumn{2}{c}{+35.27$\%$}   \\
        \bottomrule
      \end{tabular}
    }
  \end{sc}
\end{table*}
\begin{figure*}[t]
    \centering
    \includegraphics[width=\textwidth]{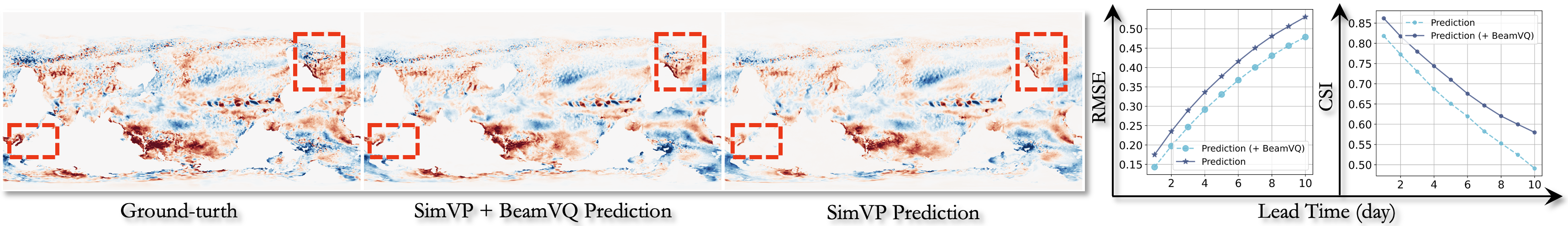}
    \caption{\textbf{The prediction results of marine extreme heatwave events include}: A visual comparison (from left to right): ground truth labels, SimVP+\method{} prediction results on day 10, and SimVP prediction results on day 10. The cumulative changes of RMSE over prediction time. The cumulative changes of CSI over prediction time.}
    \label{fig:mh_icml}
\end{figure*}
In this section, we verify the effectiveness of our method by evaluating 5 benchmarks and 10 backbone models. The experiments aim to answer the following research questions:
\textit{{\textbf{RQ1.}}} Can~\method{} enhance the performance of the baselines? 
\textit{{\textbf{RQ2.}}} How does~\method{} perform under data-scarce conditions?
\textit{\textbf{RQ3.}} Can~\method{} have better physical alignment? 
\textit{\textbf{RQ4.}} Can~\method{} produce long-term forecasting?  
Appendix~\ref{sec:more_experiments} also has additional results.

\subsection{Experimental Settings}
\textbf{Benchmarks \& Backbones.} Our dataset spans multiple spatiotemporal dynamical systems, summarized as follows: \textbf{$\bullet$ Real-world Datasets}, including SEVIR~\cite{veillette2020sevir}; \textbf{$\bullet$ Equation-driven Datasets}, focusing on PDE~\cite{takamoto2022pdebench} (Navier-Stokes equations, Shallow-Water Equations) and Rayleigh-Bénard convection flow~\cite{wang2020towards}; \textbf{(3) Computational Fluid Dynamics Simulation Datasets}, namely Prometheus~\cite{wu2024prometheus}.  We select core models from three different fields for analysis. Specifically: \textbf{$\bullet$ Spatio-temporal Predictive Learning}, we choose ResNet~\cite{he2016deep}, ConvLSTM~\cite{shi2015convolutional}, Earthformer~\cite{gao2022earthformer}, SimVP-v2~\cite{tan2022simvp}, TAU~\cite{tan2023temporal},  Earthfarseer~\cite{wu2024earthfarsser}, and FourcastNet~\cite{pathak2022fourcastnet}as representative models; \textbf{$\bullet$ Neural Operator}, we compare models like FNO~\cite{li2020fourier}, NMO~\cite{wu2024neural} and CNO~\cite{raonic2024convolutional};

\textbf{Metric.} We use Mean Squared Error (MSE) as the evaluation metric to assess each model's prediction performance. Additionally, to thoroughly evaluate the model's performance on specific tasks, we employ metrics such as Root Mean Squared Error (RMSE), Critical Success Index (CSI), Structural Similarity Index (SSIM), relative L2 error, and Turbulent Kinetic Energy (TKE). More details can be found in Appendix~\ref{sec:metric}.

\textbf{Implementation details.} Our method trains with MSE loss, uses the ADAM optimizer~\cite{kingma2014adam}, and sets the learning rate to $10^{-3}$. We set the batch size to 10. The training process early stops within 500 epochs. 
Additionally, we set our code bank size as $1024\times 64$, beam size $K$ as 5 or 10, and the threshold as the first quartile of all candidate's scores, which we find suitable for all backbones. We implement all experiments in PyTorch~\cite{paszke2019pytorch}. Training and inference for all our experiments run on a single NVIDIA A100-PCIE-40GB GPU.

\begin{table}[htbp]
  \centering
  \small
  \begin{sc}
    \caption{Comparison of backbones on marine heatwaves to evaluate \method{}'s ability to capture extreme events.}
    \label{tab:heatwaves}
      \begin{tabular}{lccc}
        \toprule
        \multirow{2}{*}{Model} & \multicolumn{2}{c}{MSE} & \multirow{2}{*}{Promotion (\%)} \\ 
        \cmidrule(lr){2-3}
        & Ori & +\method{} &  \\ 
        \midrule
        U-Net & 0.0968 & \textbf{0.0848} & 12.40\% \\
        ConvLSTM & 0.1204 & \textbf{0.0802} & \textbf{33.38\%} \\
        SimVP & 0.0924 & \textbf{0.0653} & 29.33\% \\
        \bottomrule
      \end{tabular}%
  \end{sc}
\end{table}

\begin{figure*}[t]
    \centering
    \includegraphics[width=\textwidth]{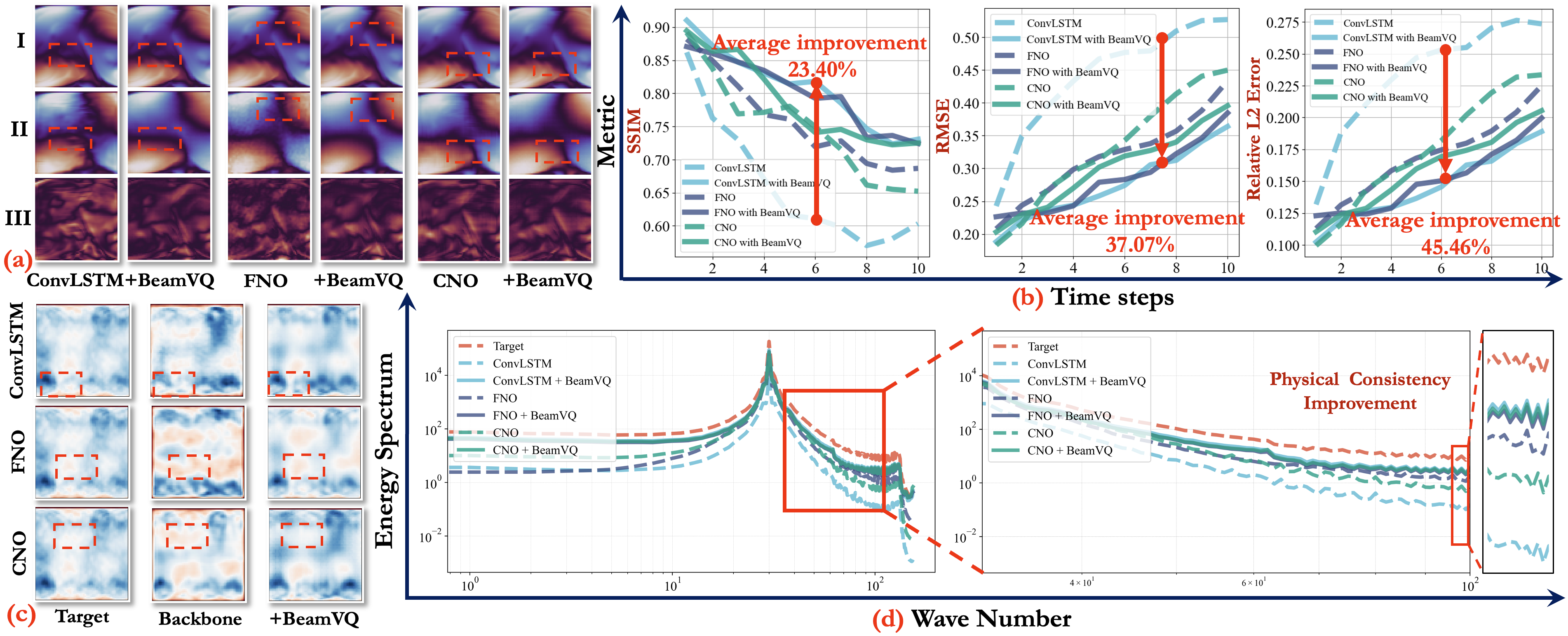}
    \caption{\textbf{The \method{} plugin improves physical consistency and prediction accuracy.} \textcolor{red}{(a)} shows a visual comparison of the actual target, predicted results, and errors at different time steps. \textcolor{red}{(b)} displays the changes in SSIM, RMSE, and relative L2 error over time steps. \textcolor{red}{(c)} compares the turbulent TKE. \textcolor{red}{(d)} presents the energy spectrum at different wavenumbers.}
    \label{fig:phy}
\end{figure*}
\begin{figure*}[h]
    \centering
    \includegraphics[width=\textwidth]{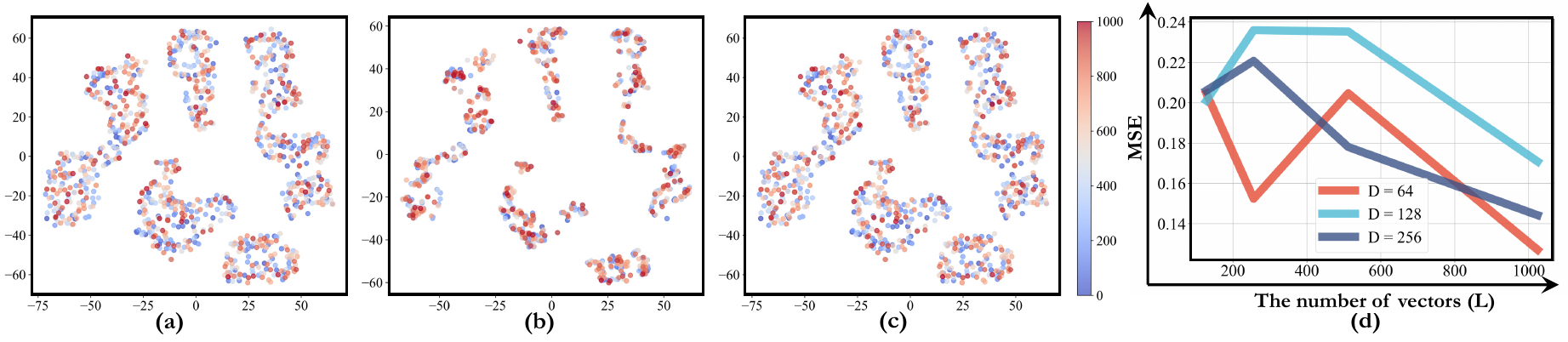}
    \caption{The t-SNE visualization in (a), (b), and (c) shows the Ground-truth, ConvLSTM and ConvLSTM+\method{} predictions, respectively. (d) shows the analysis of the Codebank parameters.}
    \label{fig:tsne} 
\end{figure*}

\subsection{\method{} improves all backbone models (RQ1)}

As shown in Table~\ref{tab:mainres}, \method{} significantly improves performance across five benchmark tests and ten backbone models. After introducing \method{}, all models show a decreasing trend in MSE, with average improvements ranging from 18.97\% to 39.08\%. For example, in the RBC fluid convection task, ConvLSTM's MSE decreases from 0.2726 to 0.0868 (a 68.15\% improvement), indicating a strong enhancement in capturing complex physical dynamics. Earthformer’s MSE in the NSE turbulence prediction task drops sharply from 1.8720 to 0.1202 (a 93.58\% improvement), demonstrating \method{}'s advantage in modeling high-dimensional chaotic systems. Even advanced models like SimVP-v2 and TAU see MSE reductions of 49.21\% and 50.85\%, respectively, in the SEVIR extreme weather prediction task, proving \method{}'s compatibility with advanced architectures. In the Prometheus combustion dynamics task, the FNO operator reduces MSE by 34.47\% (from 0.3472 to 0.2275), highlighting its enhanced ability to incorporate physical constraints. FourcastNet's MSE in the RBC task decreases from 0.0671 to 0.0219 (a 67.36\% improvement). The lightweight ResNet model's MSE in the SWE shallow water equations task drops by 56.58\% (from 0.0076 to 0.0033), demonstrating that significant accuracy gains can be achieved without complex architectures.

\subsection{\method{} helps alleviate data scarcity (RQ2)}
In scientific computing, data scarcity is a core challenge. We use extreme marine heatwaves as a scenario closely linked to human activities and economic development. To evaluate model performance, we adopt RMSE (numerical accuracy) and CSI (extreme-event capture). We compare U-Net, ConvLSTM, and SimVP as backbone networks. Table~\ref{tab:heatwaves} and Figure~\ref{fig:mh_icml} present the results, followed by our analysis. First, Table~\ref{tab:heatwaves} shows that \method{} significantly lowers MSE on the extreme marine heatwave task (e.g., ConvLSTM’s error decreases from 0.1204 to 0.0802, SimVP’s from 0.0924 to 0.0653). Even in data-scarce scenarios, these models better capture dynamic changes and improve overall prediction accuracy. Second, Figure~\ref{fig:mh_icml} compares day-10 visualizations and plots RMSE and CSI over time, indicating that \method{} generates distributions closer to the real sea temperature fields. The cumulative RMSE remains lower for models with \method{}, and CSI stays high, suggesting stronger sensitivity to extreme events and more accurate forecasts throughout the prediction period.

\subsection{\method{} Boosts Physical Alignment (RQ3)}
Figure~\ref{fig:phy} shows that \method{} significantly enhances physical consistency and prediction accuracy. In Figure~\ref{fig:phy}(a), comparing actual targets, predicted results, and errors at different time steps reveals better detail and physical consistency with smaller errors when using \method{}. Figure~\ref{fig:phy}(b) shows that \method{} improves SSIM by 23.40\%, and reduces RMSE and relative L2 error by 37.07\% and 45.46\%, respectively, indicating stronger robustness in spatiotemporal prediction. Figure~\ref{fig:phy}(c) compares turbulent kinetic energy (TKE), demonstrating more accurate capture of TKE changes, especially in small-scale turbulence. Figure~\ref{fig:phy}(d) displays the energy spectrum at different wavenumbers, where \method{} maintains better physical consistency in high-wavenumber regions—indicative of more accurate small-scale vortex prediction. Overall, \method{} not only improves numerical accuracy but also better captures the essence of physical phenomena.

\subsection{\method{} Excels In Long-term Dynamic System Forecasting (RQ4)}   
\label{sec:long}
In our long-term forecasting experiments on the SWE benchmark, we compare different backbone models by evaluating the relative L2 error for three variables (U, V, and H). We input 5 frames and predict 50 frames. For the SimVP-v2 model, using \method{} reduces the relative L2 error for SWE (u) from 0.0187 to 0.0154, SWE (v) from 0.0387 to 0.0342, and SWE (h) from 0.0443 to 0.0397, with the 3D visualization of SWE (h) shown in Figure~\ref{fig:case} [\textcolor{red}{I}]. For ConvLSTM, applying \method{} reduces the relative L2 error for SWE (u) from 0.0487 to 0.0321, SWE (v) from 0.0673 to 0.0351, and SWE (h) from 0.0762 to 0.0432. For FNO, using \method{} reduces the relative L2 error for SWE (u) from 0.0571 to 0.0502, SWE (v) from 0.0832 to 0.0653, and SWE (h) from 0.0981 to 0.0911. These results, obtained under a consistent experimental protocol, underscore the efficacy of \method{} in systematically mitigating prediction errors over extended time horizons, thereby enhancing the stability and robustness of each model's forecasts. Overall, \method{} significantly enhances the long-term forecasting accuracy of these backbone models, offering promising implications for its application in complex dynamical systems and real-world fluid dynamics scenarios.
\begin{table}[t]
  \centering
   \small
  \begin{sc}
    \caption{We compare different backbones on the SWE Benchmark for Long-term Forecasting.}
    \label{tab:time}
    \small
      \begin{tabular}{lccccc}
        \toprule
        \multirow{2}{*}{Model} & \multirow{2}{*}{} & \multicolumn{3}{c}{SWE} \\ 
        \cmidrule(lr){3-5}
        & & (u) & (v) & (h) \\ 
        \midrule
        \multirow{2}{*}{Simvp-v2} & Ori & 0.0187 & 0.0387 & 0.0443 \\
        & +\method{} & \textbf{0.0154} & \textbf{0.0342} & \textbf{0.0397} \\ 
        \midrule
        \multirow{2}{*}{ConvLSTM} & Ori & 0.0487 & 0.0673 & 0.0762 \\
        & +\method{} & \textbf{0.0321} & \textbf{0.0351} & \textbf{0.0432} \\ 
        \midrule
        \multirow{2}{*}{FNO} & Ori & 0.0571 & 0.0832 & 0.0981 \\
        & +\method{} & \textbf{0.0502} & \textbf{0.0653} & \textbf{0.0911} \\ 
        \midrule
        \multirow{2}{*}{CNO} & Ori & 0.1283 & 0.1422 & 0.1987 \\
        & +\method{} & \textbf{0.0621} & \textbf{0.0674} & \textbf{0.0965} \\ 
        \bottomrule
      \end{tabular}%
  \end{sc}
\end{table}

\begin{table}[t]
 \small
  \centering
  \begin{sc}
    \caption{Ablation studies on the NSE benchmark.}  
    \label{tab:ablation}
    \resizebox{\linewidth}{!}{%
      \begin{tabular}{l|c|c}
        \toprule
        Variants & MSE & TKE \\ 
        \midrule
        FNO & 0.2237 & 0.3964 \\
        FNO+\method{} & \textbf{0.1005} & \textbf{0.1572} \\ 
        FNO+\method{} (w/o BeamS) & 0.1207 & 0.2003 \\ 
        FNO+\method{} (w/o SelfT) & 0.1118 & 0.1872 \\ 
        FNO+\method{} (w. MSE) & 0.1654 & 0.2847 \\  
        FNO+VQVAE & 0.1872 & 0.3652 \\ 
        FNO+PINO & 0.1249 & 0.2342 \\  
        \midrule
      \end{tabular}%
    }
  \end{sc}
\end{table}

\subsection{Interpretation Analysis \& Ablation Study}

\textbf{Qualitative Analysis Using t-SNE.} Figure~\ref{fig:tsne} shows t-SNE visualizations on the RBC dataset: (a) ground truth, (b) ConvLSTM predictions, and (c) ConvLSTM + \method{} predictions. In (a), the ground truth has clear clusters. In (b), ConvLSTM’s clustering is blurry with overlaps, indicating limited capability in capturing data structure. In (c), ConvLSTM + \method{} yields clearer clusters closer to the ground truth, demonstrating that \method{} significantly enhances the model’s predictive accuracy and physical consistency.

\textbf{Analysis on Code Bank.} We train FNO+\method{} on NSE for 100 epochs with a learning rate of 0.001 and batch size of 100. In the VQVAE codebank dimension experiment, increasing the number of vectors $L$ notably reduces MSE. When $L=1024$ and $D=64$, the MSE reaches a minimum of 0.1271. Although MSE fluctuates more at $L=256$ or $512$, overall, higher $L$ helps improve accuracy. Most training losses quickly stabilize within 20 epochs; $L=512$ and $D=128$ notably shows higher stability, but $L=1024$ and $D=64$ achieves the lowest MSE.

\textbf{Ablation Study.} 
We use NSE with FNO for ablation. Variants: 
(I) \textit{FNO}; 
(II) \textit{FNO+\method{}}; 
(III) \textit{FNO+\method{} (w/o Beamsearch)}; 
(IV) \textit{FNO+\method{} (w/o self-Training)}; 
(V) \textit{FNO+\method{} (w MSE)}; 
(VI) \textit{FNO+VQVAE}; 
(VII) \textit{FNO+PINO}~\cite{10.1145/3648506}. 
Table~\ref{tab:ablation} shows FNO starts with an MSE of 0.2237 and a TKE error of 0.3964. Adding \method{} drops them to 0.1005 and 0.1572. Omitting Beamsearch or self-training increases MSE but still outperforms the base. VQVAE and PINO yield MSEs of 0.1872 and 0.1249, with TKE errors of 0.3652 and 0.2342. Overall, \method{} significantly enhances accuracy.

\section{Conclusion}
We propose BeamVQ, a unified framework for spatio-temporal forecasting in data-scarce settings. By combining VQ-VAE with beam search, BeamVQ addresses limited labeled data, captures extreme events, and maintains physical consistency. It first learns main dynamics via a deterministic base model, then encodes predictions with Top-K VQ-VAE to produce diverse, plausible outputs. A joint optimization process guided by domain-specific metrics boosts accuracy and extreme event sensitivity. During inference, beam search retains multiple candidate trajectories, balancing exploration of rare phenomena with likely system trajectories. Extensive experiments on weather and fluid dynamics tasks show improved prediction accuracy, robust extreme state detection, and strong physical consistency. Ablation studies confirm the crucial roles of vector quantization and beam search in enhancing performance.
\section*{Acknowledgements}
This work was supported by the National Natural Science Foundation of China (42125503, 42430602).

% \clearpage
\bibliography{main_arxiv}
\bibliographystyle{icml2024}

\newpage
\appendix
\onecolumn

\clearpage

\section{Metric}
\label{sec:metric}

\textbf{Mean Squared Error (MSE)} Mean Squared Error (MSE) is a common statistical metric used to assess the difference between predicted and actual values. The formula is:
\begin{equation}
    MSE = \frac{1}{n} \sum_{i=1}^{n} (y_i - \hat{y}_i)^2
\end{equation}
where $ n $ is the number of samples, $ y_i $ is the actual value, and $ \hat{y}_i $ is the predicted value.

\textbf{Relative L2 Error} Relative L2 error measures the relative difference between predicted and actual values, commonly used in time series prediction. The formula is:
\begin{equation}
    \text{Relative L2 Error} = \frac{\| Y_{\text{pred}} - Y_{\text{true}} \|_2}{\| Y_{\text{true}} \|_2}
\end{equation}
where $ Y_{\text{pred}} $ is the predicted value and $ Y_{\text{true}} $ is the actual value.

\textbf{Structural Similarity Index Measure (SSIM)} The Structural Similarity Index (SSIM) measures the similarity between two images in terms of luminance, contrast, and structure. The formula is:
\begin{equation}
    SSIM(x, y) = \frac{(2\mu_x \mu_y + C_1)(2\sigma_{xy} + C_2)}{(\mu_x^2 + \mu_y^2 + C_1)(\sigma_x^2 + \sigma_y^2 + C_2)}
\end{equation}
where $ \mu_x $ and $ \mu_y $ are the mean values, $ \sigma_x $ and $ \sigma_y $ are the standard deviations, $ \sigma_{xy} $ is the covariance.

\section{Related Work}
\label{sec:related_work}
\begin{itemize}
    \item \textbf{Numerical Methods and Ensemble Forecasting}: they are the traditional methods to realize physical spatial-temporal forecasting~\cite{jouvet2009numerical, rogallo1984numerical, orszag1974numerical, griebel1998numerical}, which employ discrete approximation techniques to solve sets of equations derived from physical laws. Although these physics-driven methods ensure compliance with fundamental principles such as conservation laws~\cite{karpatne2017theory,karnopp2012system,pukrushpan2004control}, they require highly trained professionals for development~\cite{lam2022graphcast}, incur high computational costs~\cite{pathak2022fourcastnet}, are less effective when the underlying physics is not fully known~\cite{takamoto2022pdebench}, and cannot easily improve as more observational data become available~\cite{lam2022graphcast}.
    Moreover, traditional numerical methods usually perturb initial observation inputs with different random noises, which can alleviate the problem of observation errors.  
    Then Ensemble Forecasting \cite{LEUTBECHER20083515, karlbauer2024advancing} can average the outputs of different noisy inputs to improve the robustness.
    
    \item \textbf{Data-Driven Methods}: Recently, data-driven deep learning starts to revolutionize the space of space-time forecasting for complex physical systems~\cite{gao2022earthformer, wu2024earthfarsser, li2020fourier, tan2022simvp, shi2015convolutional, pathak2022fourcastnet, wu2023pastnet,bi2023accurate,lam2022graphcast,zhang2023skilful}. Rather than relying on differential equations governed by physical laws, the data-driven approach constructs model by optimizing statistical metrics such as Mean Squared Error (MSE), using large-scale datasets. These methods~\cite{wang2022predrnn, shi2015convolutional, wang2018eidetic, tan2022simvp, gao2022earthformer, wu2024earthfarsser} are orders of magnitude faster, and excel in capturing the intricate patterns and distributions present in high-dimensional nonlinear systems~\cite{pathak2022fourcastnet}. Despite their success, purely data-driven methods fall short in generating physically plausible predictions, leading to unreliable outputs that violate critical constraints~\cite{bi2023accurate, pathak2022fourcastnet, wu2024neural}.

Previous works have tried to combine physics-driven methods and data-driven methods to get the best of both worlds. Some methods try to embed physical constraints in the neural network~\cite{long2018pde,greydanus2019hamiltonian,cranmer2020lagrangian,guen2020disentangling}. For example, PhyDNet~\cite{guen2020disentangling} adds a physics-inspired PhyCell in the recurrent network. However, such methods require explicit formulation of the physical rules along with specialized designs for network architectures or training algorithms. As a result, they lack flexibility and cannot easily adapt to different backbone architectures. Another type of methods~\cite{raissi2019physics,li2021physics,hansen2023learning}, best exemplified by the Physics-Informed Neural Network (PINN)~\cite{raissi2019physics}, leverages physical equations as additional regularizers in neural network training~\cite{hansen2023learning}. Physics-Informed Neural Operator (PINO)~\cite{li2021physics} extends the data-driven Fourier Neural Operator (FNO) to be physics-informed by adding soft regularizers in the loss function. However, PDE-based regularizers not only impose multiple-object optimization challenges~\cite{krishnapriyan2021characterizing,wang20222} but also only work in limited scenarios where has simplified physical equations and fixed boundary conditions. More recently, PreDiff~\cite{gao2024prediff} trains a latent diffusion model for probabilistic forecasting, and guides the model's sampling process with a physics-informed energy function. However, PreDiff requires training a separate knowledge alignment network to integrate the physical constraints, which is not needed in our method. 

Most importantly, all the above-mentioned works require large-scale datasets to train for good performance, while collecting scientific data can be expensive and even infeasible sometimes.
Still worse,
they can suffer from the poor prediction of extreme events, since there are sparse and imbalance extreme event data even in the large datsets.

\item \textbf{Data Augmentation:} CV has a long history of employing data augmentation to improve generalization. 
Traditionally, almost all CV works manipulate the semantic invariance in images to conduct various data pre-processing, such as random cropping, resizing, flipping, rotation, color normalization, and so on~\cite{kumar2024image}. 
More recently, CV has developed advanced techniques, such as mixup, to generate new data samples by combining different images and their labels, which can achieve even better generalization.
However, these data augmentation techniques are domain-specific, which is based on the domain knowledge of CV~\cite{kumar2024image}. 
Other fields, such as NLP, audio, and physical spatiotemporal forecasting, cannot directly adopt the same data augmentation techniques.

\item \textbf{Self-training:} it has proved to be an effective semi-supervised learning method that exploits the extra unlabeled data~\cite{du2020self}. 
Typically, self training first gives the pseudo labels on the unable data. Then it estimates the confidence of its own classification, and adds the high-confidence samples into training sets to improve the model training. 
However, our work does not have access to extra unlabeled data, making it improper to employ the existing self-training strategies.

\item \textbf{Self-ensemble:} it is well known that Ensemble methods can enhance the performance~\cite{caruana2004ensemble} and improve the robustness ~\cite{tramer2017ensemble}. 
Since Ensemble of different models needs high training costs of multiple models,
self-ensemble~\cite{wang2022self} typically makes use of different states in the training process to be free of extra training costs. 
However, existing self-ensemble cannot explore rare but critical phenomena beyond
the original dataset.

\end{itemize}

\section{Detailed Mathematical Proof}
\label{sec:proof}
\textbf{Proof of Theorem 1}

Now we have N augmented data and we need to select the best from them. We consider both the quality and the diversity of these data and get the sampling strategy from an optimization problem.

We model the sampling strategy as a multinomial distribution supported on all the augmented data $S = \{\mathbf{X}_j\}_{j=1}^N$, which means that the sampling strategy $\pi=(\pi_1,...,\pi_N)^\top$ is the corresponding probabilities of selecting $\mathbf{X}_1,...,\mathbf{X}_N$, then we can model the expectation of the similarity as:
$$\begin{aligned}
 & \mathbb{E}_{Y_x,Y_{x^{\prime}}\in\mathcal{C}}\{g(x,x^{\prime})\mid S\} \\
 & =\quad\int g(\mathbf{x},\mathbf{x}^{\prime})\boldsymbol{\pi}(\mathbf{x})\mathrm{Pr}_{S}(Y_{x}\in\mathcal{C}\mid\boldsymbol{x}=\mathbf{x})\boldsymbol{\pi}(\mathbf{x}^{\prime})\mathrm{Pr}_{S}(Y_{x}\in\mathcal{C}\mid\boldsymbol{x}=\mathbf{x}^{\prime})d\mathbf{x}d\mathbf{x}^{\prime} \\
 & =\quad\sum_{i,j=1}^Ng(\mathbf{X}_i,\mathbf{X}_j)\pi_i\pi_j\mathrm{Pr}_{S}(Y_x\in\mathcal{C}\mid\boldsymbol{x}=\mathbf{X}_i)\mathrm{Pr}_{S}(Y_x\in\mathcal{C}\mid\boldsymbol{x}=\mathbf{X}_j),
\end{aligned}$$
where the set $\mathcal{C}$ denotes the criterion of selection we are using, the function $g$ can be chosen as any similarity metric function and $x$ means a random variable.

The core to solving the above optimization problem is to use predictive inference to approximate the conditional probability of $\{Y_x\in\mathcal{C}\}$ given $x = \mathbf{X}$
Let $\mu ( \mathbf{x} ) : = \mathbb{E} ( Y\mid \mathbf{X} = \mathbf{x} )$ be the oracle associated with $( \mathbf{X} , Y) .$ Denote $\theta_j=\mathbb{I}\{Y_j\in\mathcal{C}\}$. As the augmented data
$\mathbf{X}_1,...,\mathbf{X}_N$ are independently identically distributed, $\theta_1,...,\theta_N$ can be regarded as independent Bernoulli($q)$ variables with $q=\Pr(Y_j\in\mathcal{C}).$ The probability distribution of the predicted result $W_j$ for $j=1,...,N$ is
$$\Pr(W_j\mid\theta_j)=(1-\theta_j)f_0+\theta_jf_1,\quad$$
where $f_0$ and $f_1$ are the conditional distributions of $W_j$ on $Y_j \in \mathcal{C}$ or not.

Denote $T(w) = \frac{(1-q)f_0(W_j)}{f(W_j)}$, we can rewrite the expectation of the similarity as
$$\mathbb{E}_{Y_x,Y_{x^{\prime}}\in\mathcal{C}}\{g(x,x^{\prime})|S\}=\sum_{i,j=1}^Ng(\mathbf{X}_i,\mathbf{X}_j)\pi_i\pi_j(1-T_i)(1-T_j)=\boldsymbol{\pi}^\top A_\mathbb{T}\boldsymbol{\pi},$$

Next, we use the expectation to control the quality of the data.
$$\mathbb{E}\{\mathbb{I}(Y_x\not\in\mathcal{C})\mid S\}=\sum_{i=1}^N\Pr(Y_i\not\in\mathcal{C}\mid\mathbf{X}_i)\pi_i=\sum_{i=1}^N\pi_iT_i\leq\alpha,$$

In all, the optimization problem can be modeled as 
\begin{align}
    & \arg\min_{\boldsymbol{\pi}}\quad h(\boldsymbol{\pi},\mathbb{T}):=\boldsymbol{\pi}^\top A_\mathbb{T}\boldsymbol{\pi}, \\
    & \text{subject to} \quad
        \begin{cases}
            \sum_{i = 1}^N\pi_iT_i\leq\alpha, \\
            \sum_{i = 1}^N\pi_i = 1, \\
            0\leq\pi_i\leq m^{-1}, \quad 1\leq i\leq N.
        \end{cases}
\end{align}

where $m$ is used to control the maximum selection.

The best selection of K is determined by the strategy $\pi$ which serves as the solution to the above optimization problem.

\section{Additional Experiments}
\label{sec:more_experiments}
\subsection{Long-term forecasting experiment expansion}

\begin{figure*}[h]
    \centering
    \includegraphics[width=\textwidth]{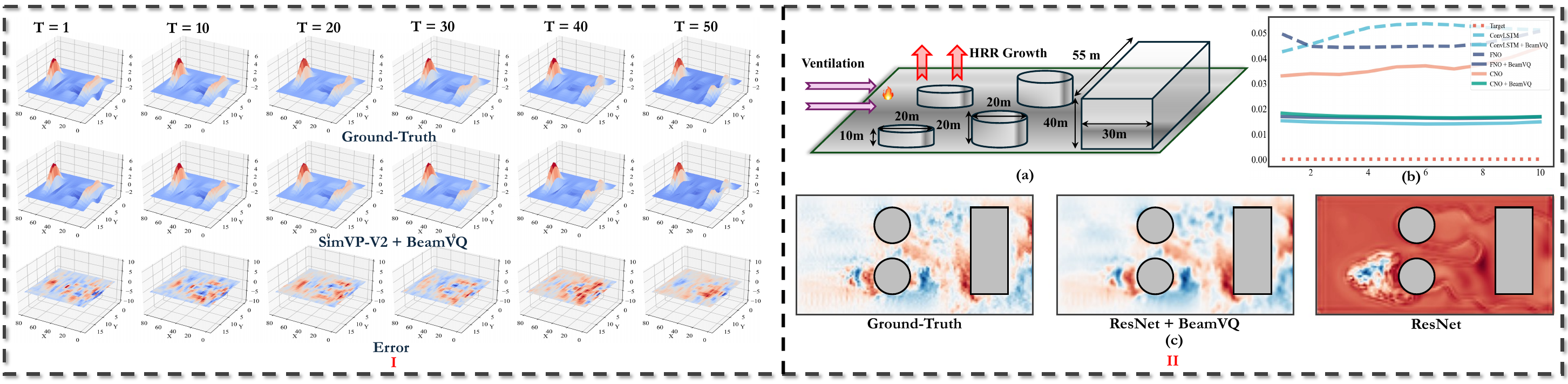}
    \caption{
    \textcolor{red}{I.} 3D visualization of the SWE(h), showing Ground-truth, SimVP-V2+BeamVQ predictions, and Error at T=1, 10, 20, 30, 40, 50. The first row shows Ground-truth, the second SimVP-V2+BeamVQ predictions, and the third Error. \textcolor{red}{II.} A case study. Building fire simulation with ventilation settings added to Wu's Prometheus~\cite{wu2024prometheus}. (a) Layout and HRR growth. (b) Comparison of physical metrics for different methods. (c) Ground-truth, ResNet+BeamVQ, and ResNet predictions.
    }
    \label{fig:case} 
\end{figure*}

In the long-term forecasting experiments, we compare the performance of different backbone models on the SWE benchmark, evaluating the relative L2 error for three variables (U, V, and H). Our setup inputs 5 frames and predicts 50 frames. For the SimVP-v2 model, using \method{} reduces the relative L2 error for SWE (u) from 0.0187 to 0.0154, SWE (v) from 0.0387 to 0.0342, and SWE (h) from 0.0443 to 0.0397. We visualize SWE (h) in 3D as shown in Figure~\ref{fig:case} [\textcolor{red}{I}]. For the ConvLSTM model, applying \method{} reduces the relative L2 error for SWE (u) from 0.0487 to 0.0321, SWE (v) from 0.0673 to 0.0351, and SWE (h) from 0.0762 to 0.0432. For the FNO model, using \method{} reduces the relative L2 error for SWE (u) from 0.0571 to 0.0502, SWE (v) from 0.0832 to 0.0653, and SWE (h) from 0.0981 to 0.0911. Overall, \method{} significantly improves the long-term forecasting accuracy of different backbone models.

\subsection{Experiment Statistical Significance}
\label{sec:significance}
To measure the statistical significance of our main experiment results, we choose three backbones to train on two datasets to run 5 times. 
Table~\ref{tab:significance} records the average and standard deviation of the test MSE loss.
The results prove that our method is statistically significant to outperform the baselines
because our confidence interval is always upper than the confidence interval of the baselines. 
Due to limited computation resources, we do not cover all ten backbones and five datasets, 
but we believe these results have shown that our method has consistent advantages.

\begin{table}[h]
\label{tab:significance}
\centering
\begin{scriptsize}
    \begin{sc}
    \caption{ The average and standard deviation of MSE in 5 runs}
    \label{tab:significance}
    \centering
        \renewcommand{\multirowsetup}{\centering}
        \setlength{\tabcolsep}{10pt}
        \begin{tabular}{l|cc|cc}
            \toprule
            
            \multirow{4}{*}{Model} & \multicolumn{4}{c}{Benchmarks}  \\
            \cmidrule(lr){2-5}
            & \multicolumn{2}{c}{NSE} &   \multicolumn{2}{c}{SEVIR}   \\
            \cmidrule(lr){2-5}
           & Ori & + BeamVQ & Ori & + BeamVQ  \\
            \midrule
            ConvLSTM &0.4092$\pm$0.0002 &\textbf{0.1277$\pm$0.0001}  & 0.1762 0.0007  & \textbf{0.1279$\pm$0.0009}  \\
            FNO &  0.2227$\pm$0.0003 &\textbf{0.1007 $\pm$0.0002}& 0.0787$\pm$0.0012 & \textbf{ 0.0437$\pm$0.0013} \\
            CNO & 0.2192 $\pm$0.0008 &\textbf{ 0.1492$\pm$0.0011}& 0.0057$\pm$0.0005 & \textbf{ 0.0053$\pm$0.0006} \\
            \bottomrule
        \end{tabular}
    \end{sc}

\end{scriptsize}
\end{table}

\end{document}